\documentclass[letterpaper, 10 pt, conference]{ieeeconf}  
\usepackage{graphicx}
\usepackage{multirow}
\usepackage{xcolor}

\usepackage{booktabs}
\usepackage{multicol}

\usepackage[colorinlistoftodos,prependcaption,textsize=tiny]{todonotes}

\graphicspath{ {./} }

\IEEEoverridecommandlockouts                              

\usepackage[a-1b]{pdfx}

\title{\LARGE \bf
Design of an End-effector with Application to Avocado Harvesting
}

\author{
{Jingzong Zhou\textsuperscript{*}\thanks{\textsuperscript{*} These authors contributed equally to this work.}, Xiao'ao Song\textsuperscript{*}} and Konstantinos Karydis
\thanks{The authors are with the Dept. of Electrical and Computer Engineering, University of California, Riverside. 
	Email: {\{jzhou227, xsong036, karydis\}@ucr.edu}. 
We gratefully acknowledge the support of NSF \# CMMI-2326309, USDA-NIFA \# 2021-67022-33453, and The University of California under grant UC-MRPI M21PR3417. Any opinions, findings, and conclusions or recommendations expressed in this material are those of the authors and do not necessarily reflect the views of the funding agencies.}
}

\begin{document}

\maketitle
\thispagestyle{empty}
\pagestyle{empty}

\begin{abstract}

Robot-assisted fruit harvesting has been a critical research direction supporting sustainable crop production. 
One important determinant of system behavior and efficiency is the end-effector that comes in direct contact with the crop during harvesting and directly affects harvesting success. 
Harvesting avocados poses unique challenges not addressed by existing end-effectors (namely, they have uneven surfaces and irregular shapes grow on thick peduncles, and have a sturdy calyx attached). 
The work reported in this paper contributes a new end-effector design suitable for avocado picking. 
A rigid system design with a two-stage rotational motion is developed, to first grasp the avocado and then detach it from its peduncle. A force analysis is conducted to determine key design parameters. 
Preliminary experiments demonstrate the efficiency of the developed end-effector to pick and apply a moment to an avocado from a specific viewpoint (as compared to pulling it directly), and in-lab experiments show that the end-effector can grasp and retrieve avocados with a $100\%$ success rate.

\end{abstract}




\section{Introduction}


The avocado tree is a high-value crop with a rapidly increasing market demand worldwide, both in terms of fresh fruit and processed products (mostly oil). 
The demand has been expanding over the years owing to the avocado's high health and nutritional values~\cite{sommaruga2021avocado, duarte2016avocado}. 
Despite the growing avocado consumption and production rates~\cite{ayala2014avocado}, the current harvesting method still relies heavily on manual hand picking with clippers and often necessitates a farmworker to climb a ladder to harvest from hard-to-reach parts of the tree \cite{mandemaker2006effects}. 
However, the increasing labor costs, coupled with labor shortages worldwide~\cite{valle2014australian, reymen2015labour}, create challenges to sustain avocado harvesting to meet market demand while remaining financially profitable for the grower~\cite{fresh_avocado_market_us}. 

One way to address these challenges (at least to a certain extent) is via the integration of agricultural robotics and automation technology in novel robot-assisted harvesting paradigms~\cite{kondo2011agricultural}. 
This paradigm involves the tight integration of multiple technological components, that is, fruit detection and localization~\cite{mccool2016visual,bac2013robust}, robot path planning~\cite{hemming2014robot,Baur2014PathPF}, and physical manipulation~\cite{bac2014harvesting}. 
Among these, one critical determinant of behavior and successful implementation is the end-effector itself; this paper focuses on end-effector design for avocado picking. 
It is worth highlighting that a recent study on using human-robot collaborative strategies during the avocado harvesting process showed that by introducing a transporting robot, the total harvested production increased from 23\% to 85\%~\cite{vasconez2022workload}. 
Yet, there is a lack of end-effectors designed for avocado harvesting to date.

\begin{figure}[!t]
\vspace{4pt}
    \centering
    \includegraphics[trim={1cm 8cm 2cm 0cm},clip,width=0.485\textwidth]{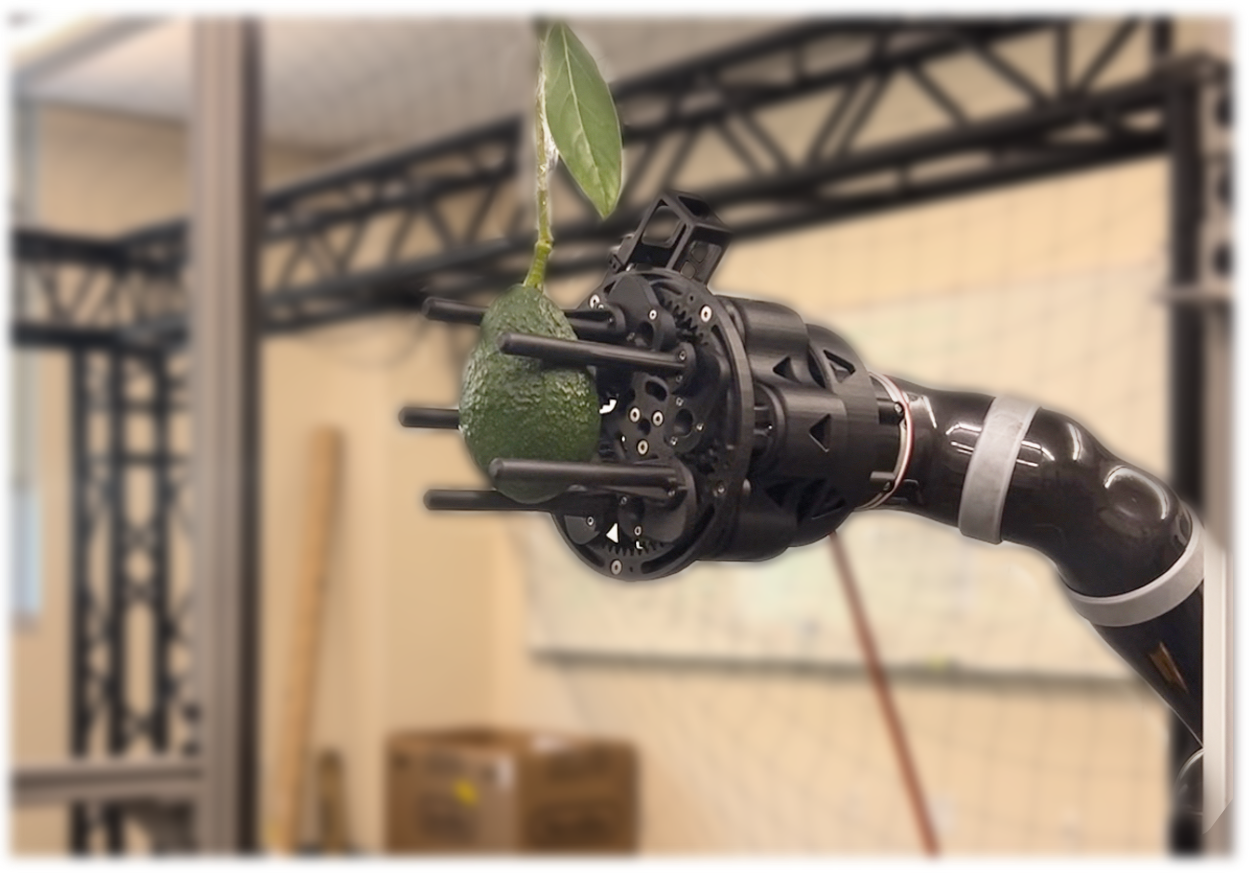}
    \vspace{-18pt}
    \caption{We developed a custom-built end-effector attached to an off-the-shelf 6-DOF robotic arm 
    for avocado harvesting.
    }
    \label{fig:gripper_demo}
    \vspace{-21pt}
\end{figure}


Several existing efforts have focused on end-effector design for fruit harvesting. 
Current designs can be broadly divided into three categories~\cite{campbell2022integrated}: vacuum-based, soft-based, and rigid-based. 
Because of their mechanical simplicity and sole requirement of coming in direct contact with the fruit, vacuum-powered suction cups are commonly used~\cite{hemming2014robot, bontsema2014crops, hayashi2010evaluation}.
Notably, Tevel, an emerging fruit harvest company, has employed this mechanism to pick up a wide range of fruits, such as apricot, apple, plum, and peach~\cite{tevel}. 
Two common features of these fruits are their near-mirror surface and a rather uniform spherical shape. 
In contrast, avocados have a less uniform shape (spheroid), and their surface is irregularly textured with bumps. 
On account of these, the effectiveness of suction cups has not yet been demonstrated. 
Soft grippers usually employ pneumatically-actuated deformable fingers that wrap around the fruit. 
They have been used mostly to pick and place light, small, fragile, and irregularly shaped fruits~\cite{navas2023soft, visentin2023soft}. 
However, soft grippers are challenged when it comes to picking heavier fruits such as avocados, and special designs to regulate how soft fingers deform are needed. 
Quite recently, an alternative method employing hybrid actuators that combine soft fingers and suction cups was proposed to pick heavier fruits like apples~\cite{wang2023development}. 
Another commonly used type of grippers includes rigid-based ones. 
They typically involve a multi-fingered design that encloses the fruit within~\cite{de2011design, ling2004sensing}, or employ a scissor-like mechanism to cut off the fruit peduncle~\cite{lehnert2017autonomous}. 
Avocados grow on thick and elastic peduncles and have a sturdy calyx attached, which makes the peduncle-cutting method hard to succeed. 
In addition, a finger-based method might be able to hold the avocado firmly but it may not be able to adapt to different avocado sizes and shapes owing to distinct avocado cultivars. 

The goal of this work is to design and test an end-effector that is suitable for harvesting avocados. 
The developed design (Fig.~\ref{fig:gripper_demo}) employs a rigid multi-fingered approach to enclose and hold an avocado. 
The main operating principle is to apply a moment to detach the avocado from its peduncle. 
To do so, we propose a rotary-based design that spins to close the fingers around the avocado and then applies an additional moment for detaching it. 
To the best of our knowledge, this is the first reported work regarding the design and testing of an end-effector specifically targeting avocados.

\section{Preliminary Experimentation for Harvesting}
\label{sec:Optimal_Avocado_Harvesting_Means}
There are several means, in general, for fruit harvesting (e.g., canopy shaking, peduncle cutting, pulling off from peduncle, or rotating the fruit along a certain axis)~\cite{navas2021soft}. 
Considering the avocado's specific characteristics (less uniform shape with irregularly textured surface with bumps, and growing on thick peduncles with sturdy calyx attached) we claim that the most efficient means to harvest an avocado is by holding the fruit and bending the peduncle so that a moment is applied on the calyx. 
In support of this argument, besides extensive empirical observations performed by the team manually picking avocados, we provide evidence from experimental testing performed to identify the amount of force required to harvest an avocado in three different ways.\footnote{~Considering that the avocado peduncle is thick and will harden when cut, it is important to remove it before packaging as it can otherwise damage other avocados in the container (e.g.,
cause punctures) and lead to faster decay of all products in the same container. To optimize the harvesting process, our goal is to thus remove the avocado from its peduncle cleanly.}

\begin{figure}[hb]
\vspace{-6pt}
    \centering    \includegraphics[width=0.40\textwidth]{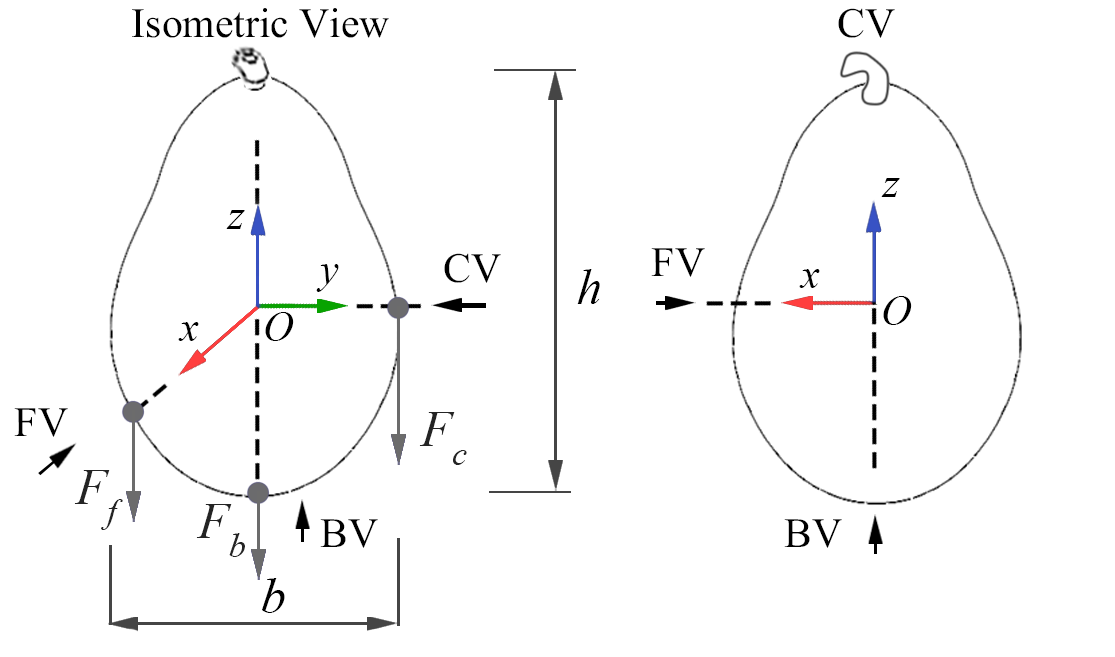}
    \vspace{-9pt}
    \caption{Fruit coordinate system and viewpoints for the avocado.}
    \vspace{-6pt}
    \label{fig: fruit_coor_system}
    \vspace{0pt}
\end{figure}
%

With reference to Fig.~\ref{fig: fruit_coor_system}, we define the viewpoint of the fruit peduncle facing toward us as the front view (FV). We also define the fruit's bottom view (BV). With these, we can then define a fruit coordinate system, centered at the middle of its centerline along the long axis. 
The viewpoint from the right side (i.e. along axis $-y$) is termed the \emph{canonical view} (CV). 
This is because the identified optimal way to harvest an avocado is by applying forces on the $x-z$ plane and a moment about the $+y$ axis.\footnote{~Due to symmetry, the left view (i.e. applying a moment about $-y$) can also be optimal; the right view was selected without any loss of generality.} We demonstrate this next. 

\begin{figure}[!t]
\vspace{4pt}
    \centering    
    \includegraphics[width=0.46\textwidth]{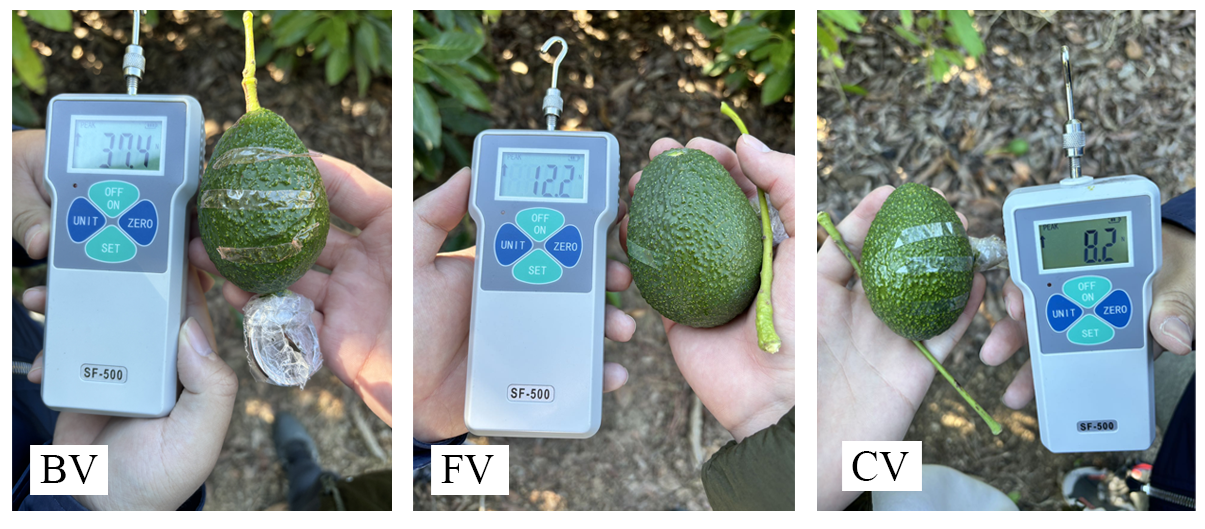}
    \vspace{-9pt}
    \caption{Instances of force measurement tests from three grasping directions.}
    \label{fig: force_test}
    \vspace{-18pt}
\end{figure}

We used a force gauge, a suspension ring, and screws for the force measurement test. 
Let the forces exerted along each direction (FV, CV, BV) be $F_f$, $F_c$ and $F_b$, respectively. Denote also the width and height of the avocado by $b$ and $h$, respectively (Fig.~\ref{fig: fruit_coor_system}). 
To measure $F_f$ or $F_c$, we inserted the screw into the avocado along the corresponding viewpoint axis. 
Then, we applied and measured a force exerted tangentially to the point of contact and which remained tangential the avocado was detached from its peduncle. 
We also tested the pull-off force ($F_b$) along the $-z$ axis direction, by inserting the screw into the bottom side of the avocado and connecting the force gauge by a suspension ring. 
Force gauge readings correspond to the minimum force needed to detach the avocado along that direction (Fig.~\ref{fig: force_test}).

\begin{table}[hb]
\vspace{-6pt}
\caption{Measured Forces on Harvested Avocados.}
\label{table: force_test}
\vspace{-9pt}
\begin{center}
\begin{tabular}{c c c c c}
\toprule
Sample No. & Viewpoint & Force [N] & $b$ [mm] & $h$ [mm]\\
\midrule
1 & \multirow{4}{*}{BV}& 22.9 & 59.52 & 75.89\\

2 &  & 41.8 & 56.70 & 78.12\\

3 &  & 37.4 & 52.75 & 75.45\\

4 &  & 26.4 & 52.85 & 78.57\\
\midrule
5 & \multirow{5}{*}{FV} & 12.2 & 62.51 & 84.58\\

6 &  & 16.9 & 61.38 & 94.83\\

7 &  & 9.6 & 60.52 & 80.84\\

8 &  & 5.6 & 50.10 & 67.83\\
\midrule
9 & \multirow{4}{*}{CV} & 10.4 & 59.78 & 90.48\\

10 &  & 11.6 & 54.78 & 68.31\\

11 &  & 8.2 & 55.61 & 77.15\\

12 &  & 8.2 & 56.66 & 80.64\\
\bottomrule

\end{tabular}
\end{center}
\vspace{-12pt}
\end{table}
We collected 12 avocados of different sizes by manually cutting them at their peduncle. We also recorded their height and width. 
Avocados were harvested randomly from various trees in the fields at the University of California, Riverside (UCR) Agricultural Experimental Station (AES). 
Avocados were fixed at their calyx to ensure consistency across the three conditions when measuring the force required to remove the avocados from their peduncle cleanly at the calyx. 


Results are shown in Table \ref{table: force_test}. 
It can be seen that pulling from the bottom is far more laborious (as in requiring considerably higher force $F_b$) than the other conditions. 
In contrast, applying a force along the canonical view to detach the peduncle at the calyx requires the least amount of force (on average). 
Considering that the branch, fruit calyx, and peduncle can all store elastic potential energy, these differences are expected to be more prevalent when harvesting directly from the tree; subsequent experiments in Section~\ref{sec:avo_grasp_exp} further support this finding.

\section{End-effector Design}
While there are a few possible means to apply a moment on the avocado along its $+y$ axis, the goal of this work is to design an end-effector that can harvest (grasp, rotate, detach, and retrieve) an avocado in one stage. 
We thus propose a multi-fingered design that first encloses and holds the avocado using a single motor and then rotates fully using a second motor to apply a moment to detach the avocado. At this stage of development, we employ the wrist joint motor of the robot arm to which the end-effector is attached. However, such a component can be readily manufactured and attached directly to the current prototype should the supporting robot arm lack a final revolute wrist joint. 

\subsection{Working Assumption}
\begin{figure}[ht]
\vspace{-12pt}
    \centering
    \includegraphics[width=0.45\textwidth]{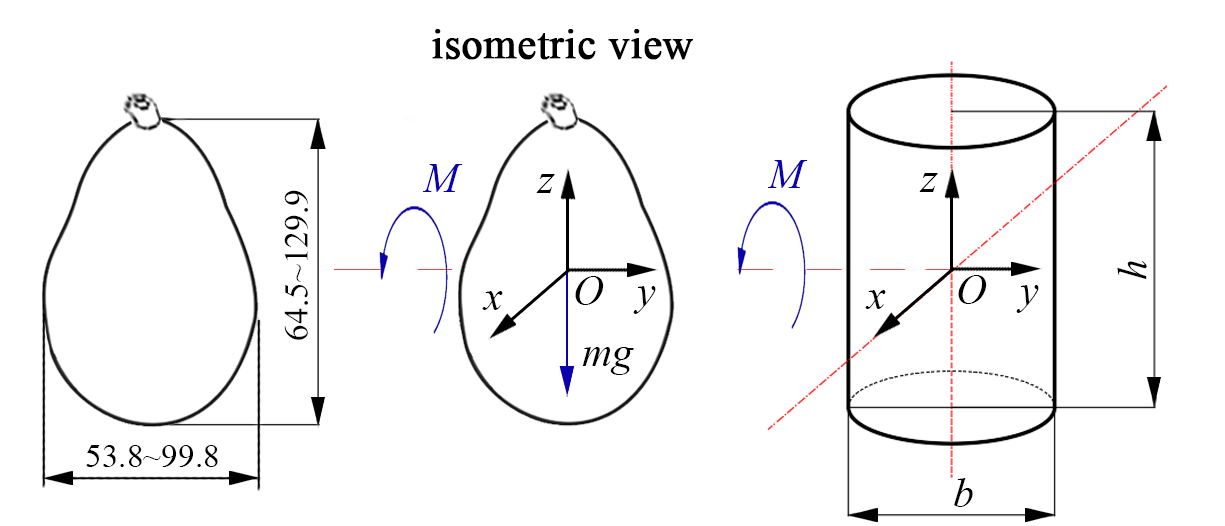}
    \vspace{-9pt}
    \caption{Our working assumption is to model an avocado as a cylinder.}
    \label{fig:avocado assumption}
    \vspace{-12pt}
\end{figure}

We model the avocado as a cylinder to expedite the force analysis required for the design and component selection of the end-effector (Fig.~\ref{fig:avocado assumption}). 
The average height and width of different avocado cultivars are in the range $[64.5,129.9]$\;mm and $[53.8,99.8]$\;mm, respectively~\cite{mokria2022volume}. 
We use the upper values to compute 1) the maximum opening the gripper should achieve, and 2) the maximum moment of inertia which in turn is used for component selection (crucially, a motor to generate sufficient torque to detach the avocado).

\subsection{Design Overview}
\begin{figure}[!t]
\vspace{4pt}
    \centering
    \includegraphics[width=0.45\textwidth]{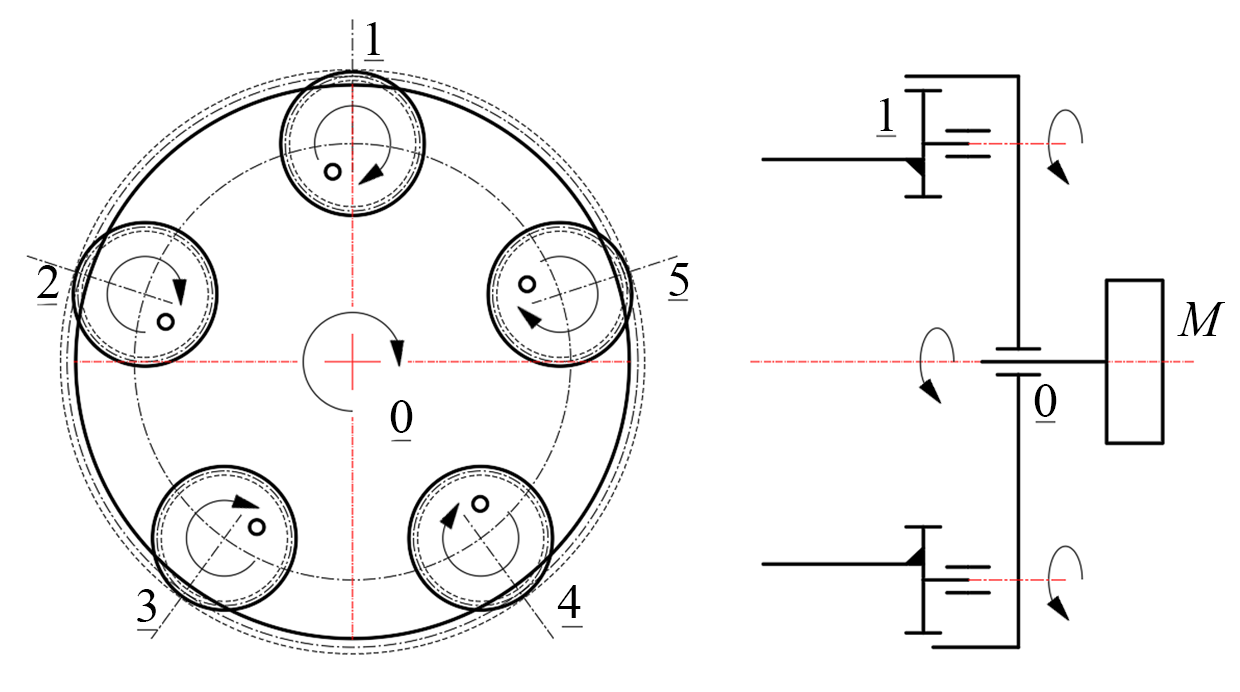}
    \vspace{-9pt}
    \caption{Mechanical diagram of the developed end-effector.}
    \label{fig:transmission diagram}
    \vspace{-3pt}
\end{figure}

The main transmission chain of the designed mechanism is depicted in Fig.~\ref{fig:transmission diagram}.  
A rotary motor denoted by $M$ impels the large internal gear 0, subsequently propelling the small external gears 1 to 5. Each small gear is affixed to a finger structure. 
Upon clockwise rotation of the internal gear, the associated finger retracts inward, facilitating a grasping motion to securely hold the target avocado. 

The finger structure is in a shape of a slender cylinder with one end attached on the small gear and the other end unsupported. 
This arrangement can be conceptualized as a cantilever. 
When the unsupported end is loaded by the reaction force from the grabbed object, the largest bending moment occurs at the gear side. 
A design comprising only two fingers is insufficient to adequately hold the object. 
Moreover, employing three or four fingers may result in an undesirable increase in the inter-finger spacing, potentially leading to unintentional release of the object through the widened gaps. 
To mitigate the risk of any single gear bearing excessive loading during the grasping process, five fingers with five small gears were used here to perform the grasping task. 
This assembly is responsible for enclosing and securing the avocado within. 
The whole assembly then rotates about its center axis to detach the avocado. 
In the current design, we attach the end-effector in a commercial off-the-shelf manipulator with a rotary wrist, in order to minimize end-effector complexity and make it more easily generalizable with complementary hardware.

\begin{figure}[!t]
\vspace{-6pt}
    \centering
    \includegraphics[width=0.45\textwidth]{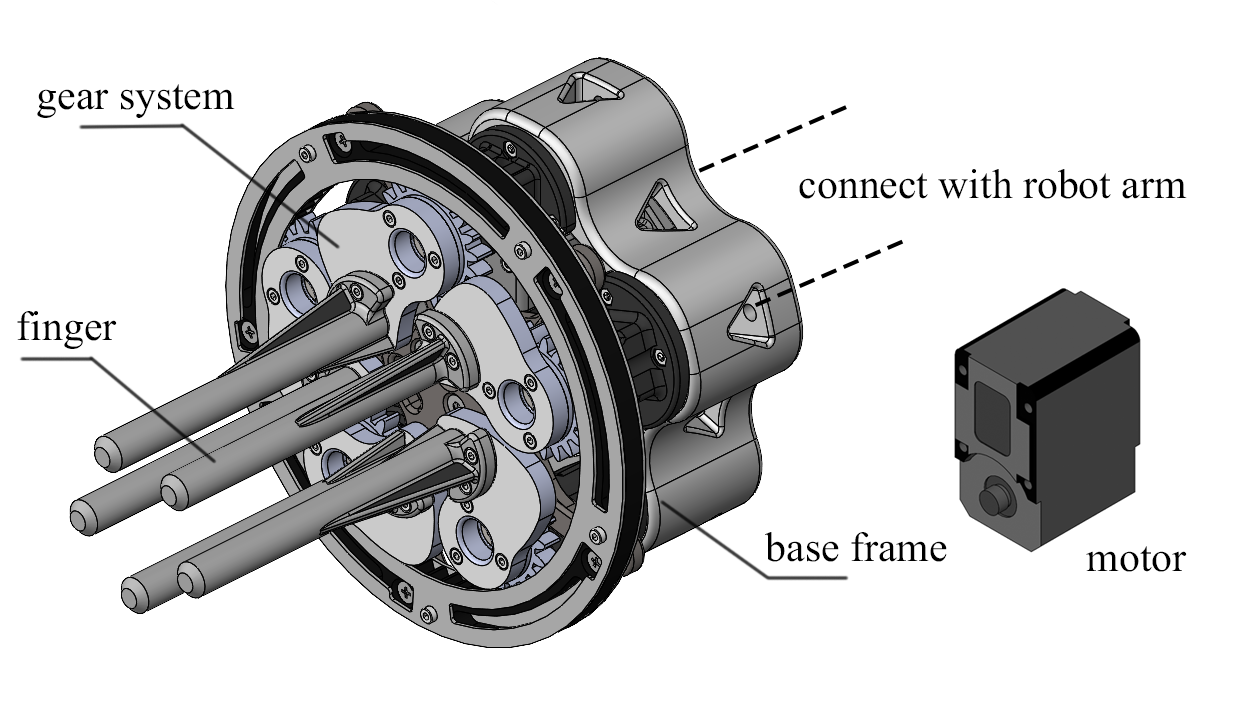}
    \vspace{-9pt}
    \caption{Mechanical design of the gripper.}
    \label{fig: mechanical design model}
    \vspace{-19pt}
\end{figure}

The end-effector prototype is depicted in Fig.~\ref{fig: mechanical design model}, with the physical prototype shown in Fig.~\ref{fig:gripper_demo}. It has three main components: rigid fingers, a base frame, and the gear system. A rotary motor is connected within the base frame to drive the gear system. 
The other end of the base frame is connected to the robot arm. 
On account of the structure of the finger-gear system, the rotary motor can only rotate in a certain range, here within $[13,110]^{\circ}$. Note that the physical prototype built by a 3D printer using carbon-fiber-reinforced material includes a rib structure not depicted in the mechanical diagram. 
Such a design helps guarantee part strength in practice. 
The force analysis excludes this feature, though a more detailed analysis can be part of future work.

\subsection{Force Analysis}
\begin{figure}[!t]
\vspace{4pt}
    \centering
    \includegraphics[width=0.45\textwidth]{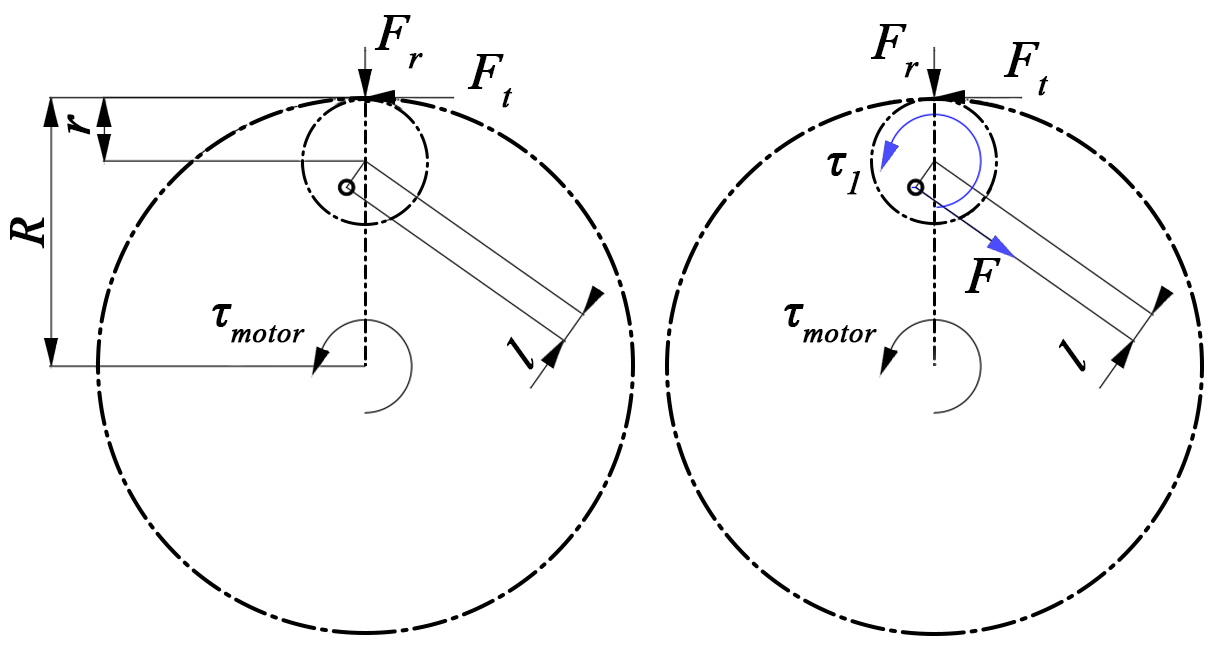}
    \vspace{-9pt}
    \caption{Force analysis on one finger of the end-effector.}
    \label{fig:gear force}
    \vspace{-6pt}
\end{figure}

Given the torque provided by the rotary motor, $\tau_M$, we seek to derive an expression for the moment relating to the interaction of five fingers engaged with all gears. 
First, we derive the force $F$ on each finger which contains two components: $F_t$, the tangent force along the tangent line of the pitch circle (circle where small gears and large gears engage); and $F_r$, the radial force along the radius of pitch circle (Fig.~\ref{fig:gear force}). 
The tangent force is $F_t = \tau_M / R$, where $R$ is the radius of large internal gear 0, and $\tau_M$ is the motor torque. 
$F_r$ is the radial force pointing to the center of the pitch circle and therefore it does not contribute any moment. 
The torque $\tau_1$ provided by $F_t$ can be computed as $\tau_1 = F_t r$, which then contributes to the force $F$ on the finger as $F = \tau_1 / l$, where $l$ is the offset distance from the center of the finger to the center of the small gear 1.

\begin{figure}[!ht]
\vspace{-6pt}
    \centering
    \includegraphics[width=0.45\textwidth]{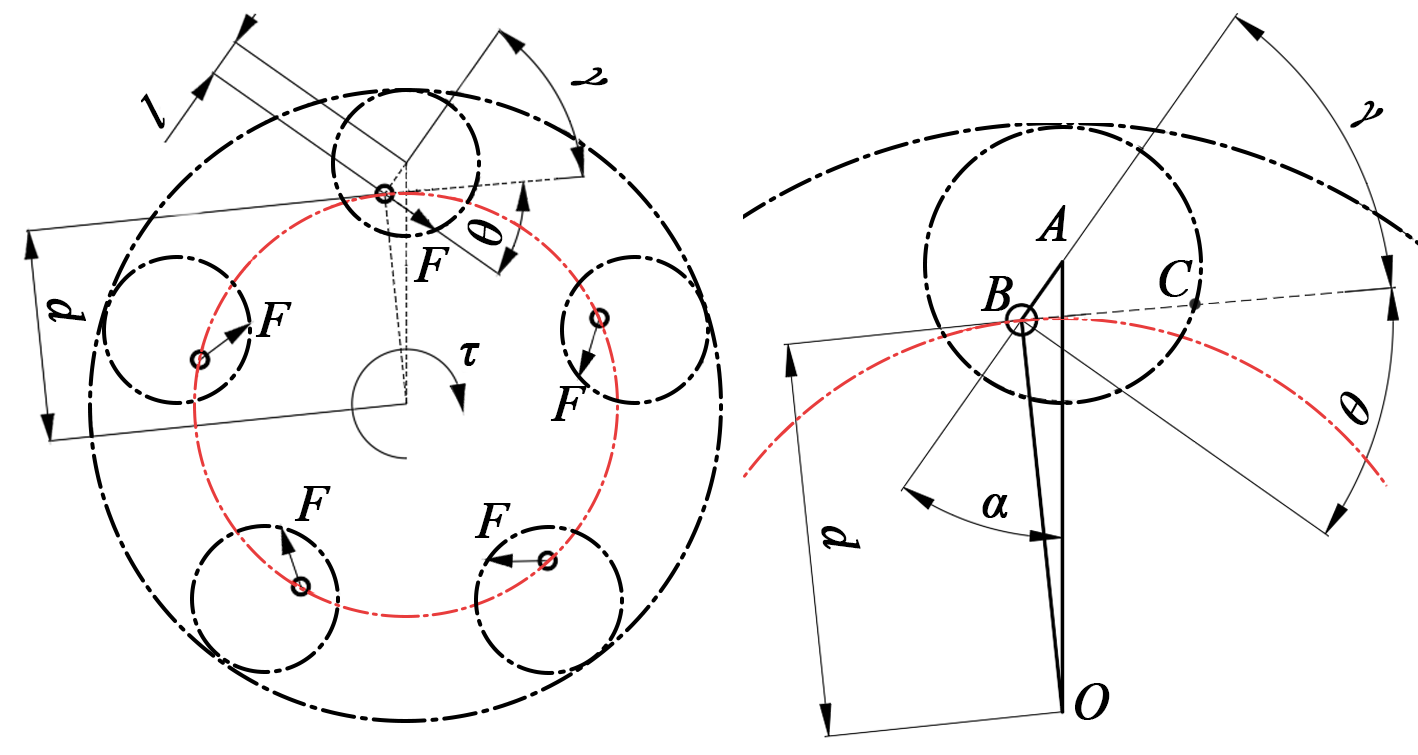}
    \vspace{-9pt}
    \caption{Force analysis on the five fingers of the end-effector.}
    \label{fig:force together}
    \vspace{-6pt}
\end{figure}


We can then derive the moment generated from all five fingers. With reference to Fig.~\ref{fig:force together}, the total moment is $\tau = 5F \cos\theta d$, where $d$ is the distance from the center of the finger to the center of the large internal gear, and $\theta$ is the angle between the force $F$ and the tangent line passing the circle formed by five centers of the finger. 
To derive the expression for angle $\theta$ we employ angles $\alpha$ and $\gamma$ (Fig.~\ref{fig:force together}). 
Angle $\alpha$ can be directly measured via the motor encoder. 
Then, from trigonometry in triangle $ABO$, the length $d$ is $d = \sqrt{l_{AB}^2 + l_{OA}^2 - 2l_{AB} l_{OA} \cos\alpha}$, where $A$ is the center of small gear, $B$ is the center of finger, $O$ is the center of large internal gear, $l_{AB}$ is the offset distance from center $A$ to $B$, and $l_{OA}$ is the distance between the center of large internal gear and the center of small gear. 
The angle $\angle ABO$ can be derived as
    $\angle ABO = asin((l_{OA} \sin\alpha) / d)$,
which in turn enables calculation of angle $\gamma$ as $\gamma = \angle ABO - \pi/2$. 
Finally, angle $\theta$ can be computed as $\theta = \pi/2 - \gamma$. 
This way, the moment, $\tau$, provided by the five fingers can be expressed with respect to angle $\alpha$ that is directly measurable.

\section{Experimental Testing and Results}
\label{sec:avo_grasp_exp}
\subsection{Experimental Setup}
We conducted the experiments in an indoor setting. The end-effector was controlled using a GEEKOM Mini PC, with DC converter and LX-16A bus servo. 
The end-effector was mounted on the last (revolute) joint of a 6-DOF Kinova Jaco2 arm to complete the rotational motion for final fruit detachment. 
We collected from the field 30 fresh Hass avocados with a long part of their peduncle attached, brought them into the lab, and numbered and labeled them in three groups evenly: small, medium, and large. 
Half of the avocados were tested with FV grasping, and the other half with CV grasping. 
Note that these avocados were still unripe with a tough surface which is not easy to deform (which matches the actual harvesting). 
Avocado sizes are shown in 
Tables~\ref{table: grasping_experiment_fv} (FV grasping) and ~\ref{table: grasping_experiment_cv} (CV grasping).

\begin{figure}[!t]
\vspace{4pt}
    \centering
    \includegraphics[width=0.45\textwidth]{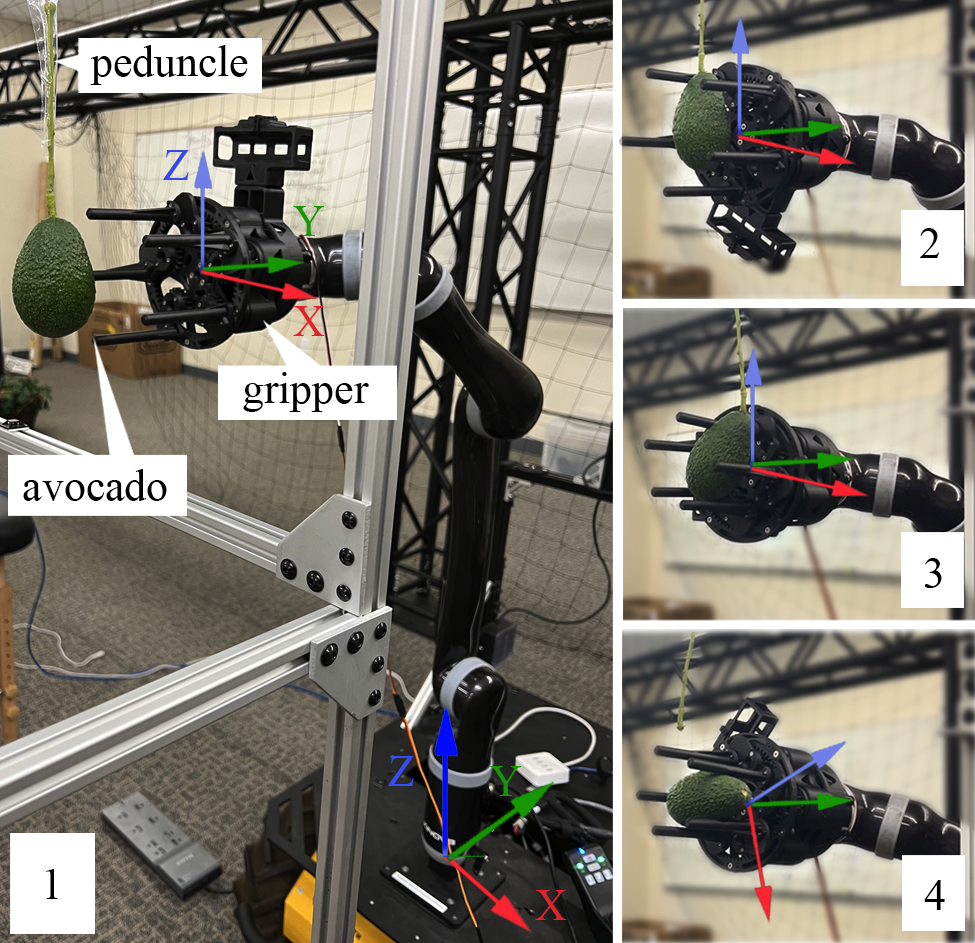}
    \vspace{-9pt}
    \caption{Grasping experimental setup in in-lab settings.}
    \label{fig: gripper_indoors_experiment}
    \vspace{-18pt}
\end{figure}

The cut tip of the peduncle was fixed on a solid frame by tape. 
The avocado was hung in its natural position to match its common pose in the orchard. 
The whole setup was placed within the manipulator's workspace. 
The world frame was attached to the base of the arm (left panel of Fig.~\ref{fig: gripper_indoors_experiment}). 
Similarly to other related works~\cite{campbell2022integrated}, we first determined a staging 6D pose for the end-effector's frame (herein we set $[-0.09,-0.53,0.84, 90.1, 5.4, 0]^T$ where the first three numbers are the 3D position coordinates in meters and the last three correspond to the orientation in Euler-XYZ angles in degrees). 
Arm motion was controlled by Jaco2 Kinova's built-in inverse kinematics solver to handle singularities, self-collision avoidance, and trajectory execution. 
The complete end-effector assembly will then move locally to engulf the avocado and activate to grasp and detach it. 

\subsection{Harvesting Work Flow}
\label{subsec:harvest_workflow}
The harvesting workflow can be divided into four processes: staging, attaching, grasping, and detaching (Fig.~\ref{fig: gripper_indoors_experiment} panels \{1,2,3,4\}, respectively). 
First, the Jaco2 arm moves from its home position to align the end-effector frame with the desired staging pose. 
Second, the end-effector is in OPEN state and then moves closer to attach the avocado within its fingers. 
Third, the end-effector is in CLOSE state to grasp the avocado by a certain preset closeness. 
Last, the rotary joint on the Jaco2 robot arm rotates to detach the avocado from its peduncle. 
This process was done manually with the Joystick, with no visual image processing involved at this stage of development (but which is part of ongoing work). 


\subsection{System Validation, Comparative Assessment of Different Grasping Poses, and Discussion of Key Findings}

The purpose of the main experimental phase is two-fold: 1) to assess the overall avocado harvesting success rate (from either FV or CV), and 2) to determine any subtle differences in the degree of actuation between the two grasping poses. 
Each time there was only one avocado being experimented with, and hence potential impact of adjacent avocados in practical deployment was not considered as part of this work. 
We considered both CV and FV because in large and irregularly shaped avocados the moment arm exerted by CV on the calyx is slightly larger than the moment arm exerted by FV, thus resulting in a smaller angle. 
When the avocado is more symmetric, this difference is less significant. 
The arm wrist angular velocity was the same, at $0.326$\;rad/s (i.e. the manufacturer's default value). 
Results (presented in Tables~\ref{table: grasping_experiment_fv} and~\ref{table: grasping_experiment_cv}) indicate the minimum degree of rotational motion required from the arm's last revolute joint to detach the avocado from its peduncle from the picking up pose.

\begin{table}[hb]
\vspace{-3pt}
\caption{Avocado Grasping Experimental Results (from FV)}

\label{table: grasping_experiment_fv}
\vspace{-12pt}
\begin{center}
\begin{tabular}{c c c c c c}
\hline
Sample No. &Group &$b$ [mm] & $h$ [mm]  & FV [$^{\circ}$]  \\
\midrule
1 & \multirow{6}{*}{Small}&48.16 & 74.36 &  80
\\ 
2 && 47.26 & 67.97 &  80\\ 
3 && 44.71& 64.33& 75\\ 
4 && 43.08 & 64.66 &  85\\ 
5 && 44.36 & 68.73 &  80\\ 
Avg. && \textbf{45.51} & \textbf{68.01} &   \textbf{80.00}  \\ 

\midrule

6 &  \multirow{5}{*}{Medium}&57.88 & 75.80 &  80 \\ 
7 & &54.59 & 76.28 &  75\\ 
8 & &56.44 & 81.53 &  80\\ 
9 & &54.61 & 81.03 & 75\\ 
10 & & 53.47 & 75.14 &  75\\ %
Avg. && \textbf{55.40} & \textbf{77.96} &   \textbf{77.00}  \\ 
\midrule
11 &  \multirow{5}{*}{Large}&64.02 & 86.64 & 90\\ %
12 & &60.67 & 89.60 & 75\\ %
13 & &57.65 & 89.71 &  85 \\ %
14 & &63.14 & 86.07 &  80\\ %
15 & &65.53 & 94.22 & 90\\ 
Avg. && \textbf{62.20} & \textbf{89.25} &   \textbf{84.00}   \\ 
\hline
\end{tabular}
\end{center}
\vspace{-12pt}
\end{table}

\begin{table}[ht]
\vspace{3pt}
\caption{Avocado Grasping Experimental Results (from CV)}
\label{table: grasping_experiment_cv}
\vspace{-12pt}
\begin{center}
\begin{tabular}{c c c c c c}
\hline
Sample No. &Group &$b$ [mm] & $h$ [mm]  & CV [$^{\circ}$]  \\
\midrule
1 & \multirow{6}{*}{Small}&47.94 & 67.96 &  70
\\ 
2 && 41.23 & 61.61&  65\\ 
3 && 44.45& 64.03& 75\\ 
4 && 43.19 & 57.25 &  80\\ 
5 && 45.51 & 63.36 &  70\\ 
Avg. && \textbf{44.46} & \textbf{62.84} &   \textbf{72.00}  \\ 

\midrule

6 &  \multirow{5}{*}{Medium}&56.30 & 74.36 &  65 \\ 
7 & &49.74 & 75.08 &  75\\ 
8 & &56.20 & 73.95 &  60\\ 
9 & &60.56 & 76.99 & 90\\ 
10 & & 55.23 & 86.56 &  65\\ %
Avg. && \textbf{55.61} & \textbf{77.39} &   \textbf{72.00}  \\ 
\midrule
11 &  \multirow{5}{*}{Large}&64.94 & 91.45 & 55\\ %
12 & &54.42 & 98.09 & 55\\ %
13 & &60.22 & 91.00 &  75 \\ %
14 & &62.47 & 86.49 &  50\\ %
15 & &61.92 & 87.10 & 65\\ 
Avg. && \textbf{60.79} & \textbf{90.83} &   \textbf{60.00}   \\ 
\hline
\end{tabular}
\end{center}
\vspace{-15pt}
\end{table}

Results demonstrate that the developed end-effector has a $100\%$ success rate in avocado harvesting in the benchmark in-lab experimental setup. 
This holds across different avocado sizes (with width and height ranging in $[41.23,65.53]$\;mm and $[57.25, 98.09]$\;mm, respectively) irrespective of the attempted grasping pose (FV/CV). 
In some cases, large deformations of the fingers may occur; however, the end-effector can still grasp the avocado. 
Results show the end-effector's adaptability to different fruit sizes, which, for the specific case of avocado is crucial, as different cultivars can vary drastically in their size and volume~\cite{mokria2022volume}. 

Further, we can observe that across all avocado size groups, grasping from the CV pose requires a smaller amount of arm-wrist rotation as compared to grasping from the FV pose. 
This result is particularly evident in the large avocado group: the average rotation angle required in CV pose is only 71.4\% of FV pose. 
This is important because it provides an instance of actuation speed and efficiency (i.e. if both CV and FV grasping poses are possible, CV is faster and thus requires less energy). 
One limiting factor of current agricultural robotics is actuation speed~\cite{campbell2022integrated}, so even small differences at a single-component experiment level may yield improved efficiency when deployed at a field scale.

\subsection{Comparison with Vacuum Gripper}
A claim made earlier was regarding the suitability of vacuum suction cup grippers. 
To validate this claim, we mounted a commercial vacuum gripper onto the Jaco2 arm and performed experiments (10 times per avocado from each size group) to test the attachment success rates.

\begin{figure}[ht]
\vspace{-9pt}
    \centering
    \includegraphics[width=0.4\textwidth]{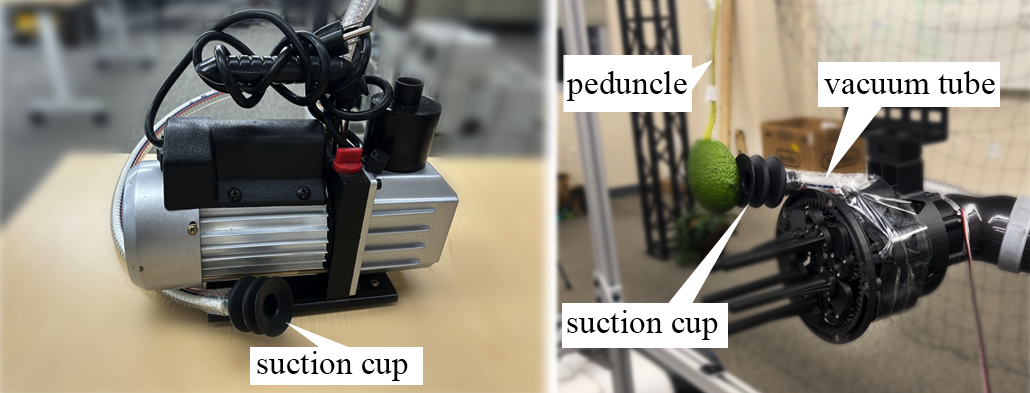}
    \vspace{-9pt}
    \caption{Vacuum gripper testing setup}
    \label{fig: vacuum_gripper_test}
    \vspace{-9pt}
\end{figure}

The setup included a $1.68$\;cm diameter soft vacuum tube and a $4.45$\;cm diameter silicone suction cup, powered by a single stage vacuum pump which can provide ultimate vacuum at $5$\;Pa (Fig.~\ref{fig: vacuum_gripper_test}). 
This vacuum gripper setup can sustain a suction force of approximately $F_{suction}=28.60$\;N as per $F_{suction} = \Delta P * A$, where $A$ is the area of the vacuum tube and $\Delta P$ is the difference between standard atmospheric pressure (set as $101,325$\;Pa) and the ultimate vacuum. 
Given that a mature Hass avocado typically weighs $0.2-0.3$\;kg, this suction force suffices to pull about 10-15 such avocados. (See supplemental video for the experimental setup.)

Results validate our initial assessment that a vacuum suction cup-based gripper would not be appropriate in the context of avocado grasping. 
We observed that the suction cup gripper could attach occasionally to smaller avocados and that only after multiple manual efforts to find a strong contact between the avocado and the suction cup. 
We also conducted an experiment where the vacuum gripper was first manually attached to a small avocado and then rotated to test if the avocado could be removed from its peduncle (assuming that the suction cup could work). 
Our findings demonstrated that the gripper would detach from the avocado at $80^{\circ}$ regardless of the pose of the suction cup with reference to the avocado. (See results in the supplemental video.)

The uneven surface of the avocado led to ambient air intake inside the suction cup and prevented attachment. 
Moreover, we tried to use a soft finger structure (made of Formlabs Form 3 Flex-80A resin) in place of the rigid one, but it was impossible to hold the avocado tightly. 
We note that a hybrid design with fingers and a suction cup (like in related work for picking apples~\cite{wang2023development}) may be possible.  However, this would come at a higher mechanism design and control operation complexity, whereas our proposed end-effector affords direct and intuitive operation.


\section{Conclusions and Future Research Directions}






Robot-assisted fruit harvesting can help enable more sustainable crop production. 
One important determinant of system behavior and efficiency is the end-effector that comes in direct contact with the crop during harvesting. 
Despite current ongoing efforts to design end-effectors across different types of crops, end-effectors for harvesting avocados--a crop with significant profit margins to the grower and health benefits to the consumers--were missing. 

In this work we developed one such end-effector. 
Our approach employed a rigid system design with a two-stage rotational motion to first grasp the avocado and then detach it from its peduncle. 
Preliminary experiments assessed the efficiency of our proposed approach to pick and apply a moment to an avocado from a specific viewpoint (as compared to pulling it directly), while more extensive in-lab experiments demonstrated a $100\%$ success harvesting rate. 
Experiments showed that vacuum suction cup-based methods can be challenged to establish and hold appropriate contact due to the avocado's uneven surface. 

The work herein creates several directions for future research. 
One important aspect is the integration of perception in static (in-lab) settings and follow-on merging with planning for fully autonomous behavior, as in our earlier work for autonomous leaf cutting~\cite{dechemi2023robotic}. 
Future work will also focus on field deployment and testing in local experimental farms. 
Additionally, a force sensor will be affixed to the design to determine appropriate grasping firmness. Further, a study of the number of fingers that can hold the avocado tight is also worthwhile as a future direction. 
Finally, we seek to conduct a larger study to identify any statistically significant differences when the avocado size varies and identify any possible links with the level of maturity/ripeness of the harvested avocado.

\addtolength{\textheight}{-12cm}   





\bibliographystyle{IEEEtran}
\bibliography{references}

\begin{thebibliography}{10}
\providecommand{\url}[1]{#1}
\csname url@samestyle\endcsname
\providecommand{\newblock}{\relax}
\providecommand{\bibinfo}[2]{#2}
\providecommand{\BIBentrySTDinterwordspacing}{\spaceskip=0pt\relax}
\providecommand{\BIBentryALTinterwordstretchfactor}{4}
\providecommand{\BIBentryALTinterwordspacing}{\spaceskip=\fontdimen2\font plus
\BIBentryALTinterwordstretchfactor\fontdimen3\font minus \fontdimen4\font\relax}
\providecommand{\BIBforeignlanguage}[2]{{%
\expandafter\ifx\csname l@#1\endcsname\relax
\typeout{** WARNING: IEEEtran.bst: No hyphenation pattern has been}%
\typeout{** loaded for the language `#1'. Using the pattern for}%
\typeout{** the default language instead.}%
\else
\language=\csname l@#1\endcsname
\fi
#2}}
\providecommand{\BIBdecl}{\relax}
\BIBdecl

\bibitem{sommaruga2021avocado}
R.~Sommaruga and H.~M. Eldridge, ``Avocado production: Water footprint and socio-economic implications,'' \emph{EuroChoices}, vol.~20, no.~2, pp. 48--53, 2021.

\bibitem{duarte2016avocado}
P.~F. Duarte, M.~A. Chaves, C.~D. Borges, and C.~R.~B. Mendon{\c{c}}a, ``Avocado: characteristics, health benefits and uses,'' \emph{Ci{\^e}ncia Rural}, vol.~46, pp. 747--754, 2016.

\bibitem{ayala2014avocado}
T.~Ayala~Silva and N.~Ledesma, ``Avocado history, biodiversity and production,'' \emph{Sustainable Horticultural Systems: Issues, Technology and Innovation}, pp. 157--205, 2014.

\bibitem{mandemaker2006effects}
A.~Mandemaker, T.~Elmsly, and D.~Smith, ``Effects of drop heights and fruit harvesting methods on the quality of'hass' avocados,'' \emph{New Zealand Avocado Growers’ Association Annual Research Report}, vol.~6, pp. 97--104, 2006.

\bibitem{valle2014australian}
H.~Valle, T.~Caboche, M.~Lubulwa \emph{et~al.}, ``Australian vegetable growing farms: an economic survey, 2011-12 and 2012-13,'' \emph{ABARES Research Report}, no. 14.01, 2014.

\bibitem{reymen2015labour}
D.~Reymen, M.~Gerard, P.~De~Beer, A.~Meierkord, M.~Paskov, V.~di~Stasio, V.~Donlevy, I.~Atkinson, A.~Makulec, U.~Famira-M{\"u}hlberger \emph{et~al.}, ``Labour market shortages in the european union,'' \emph{WIFO Studies}, 2015.

\bibitem{fresh_avocado_market_us}
\BIBentryALTinterwordspacing
O.~Bastida. (2023) Current situation of the fresh avocado market in the us. [Online]. Available: \url{https://producepay.com/blog/fresh-avocado-market-us/}
\BIBentrySTDinterwordspacing

\bibitem{kondo2011agricultural}
N.~Kondo, M.~Monta, and N.~Noguchi, \emph{Agricultural robots: mechanisms and practice}.\hskip 1em plus 0.5em minus 0.4em\relax Apollo Books, 2011.

\bibitem{mccool2016visual}
C.~McCool, I.~Sa, F.~Dayoub, C.~Lehnert, T.~Perez, and B.~Upcroft, ``Visual detection of occluded crop: For automated harvesting,'' in \emph{IEEE International Conference on Robotics and Automation (ICRA)}, 2016, pp. 2506--2512.

\bibitem{bac2013robust}
C.~Bac, J.~Hemming, and E.~Van~Henten, ``Robust pixel-based classification of obstacles for robotic harvesting of sweet-pepper,'' \emph{Computers and Electronics in Agriculture}, vol.~96, pp. 148--162, 2013.

\bibitem{hemming2014robot}
J.~Hemming, C.~W. Bac, B.~A. van Tuijl, R.~Barth, J.~Bontsema, E.~Pekkeriet, and E.~Van~Henten, ``A robot for harvesting sweet-pepper in greenhouses,'' \emph{International Conference of Agricultural Engineering}, 2014.

\bibitem{Baur2014PathPF}
\BIBentryALTinterwordspacing
J.~Baur, C.~Schuetz, J.~Pfaff, T.~Buschmann, and H.~Ulbrich, ``Path planning for a fruit picking manipulator,'' in \emph{Agricultural and Food Sciences, Engineering}, 2014. [Online]. Available: \url{https://api.semanticscholar.org/CorpusID:57513114}
\BIBentrySTDinterwordspacing

\bibitem{bac2014harvesting}
C.~W. Bac, E.~J. Van~Henten, J.~Hemming, and Y.~Edan, ``Harvesting robots for high-value crops: State-of-the-art review and challenges ahead,'' \emph{Journal of Field Robotics}, vol.~31, no.~6, pp. 888--911, 2014.

\bibitem{vasconez2022workload}
J.~P. V{\'a}sconez and F.~A.~A. Cheein, ``Workload and production assessment in the avocado harvesting process using human-robot collaborative strategies,'' \emph{Biosystems Engineering}, vol. 223, pp. 56--77, 2022.

\bibitem{campbell2022integrated}
M.~Campbell, A.~Dechemi, and K.~Karydis, ``An integrated actuation-perception framework for robotic leaf retrieval: Detection, localization, and cutting,'' in \emph{IEEE/RSJ International Conference on Intelligent Robots and Systems (IROS)}, 2022, pp. 9210--9216.

\bibitem{bontsema2014crops}
J.~Bontsema, J.~Hemming, and E.~Pekkeriet, ``Crops: high tech agricultural robots,'' \emph{International Conference of Agricultural Engineering}, 2014.

\bibitem{hayashi2010evaluation}
S.~Hayashi, K.~Shigematsu, S.~Yamamoto, K.~Kobayashi, Y.~Kohno, J.~Kamata, and M.~Kurita, ``Evaluation of a strawberry-harvesting robot in a field test,'' \emph{Biosystems Engineering}, vol. 105, no.~2, pp. 160--171, 2010.

\bibitem{tevel}
\BIBentryALTinterwordspacing
Tevel-tech. (2022). [Online]. Available: \url{https://www.tevel-tech.com/}
\BIBentrySTDinterwordspacing

\bibitem{navas2023soft}
E.~Navas, R.~R. Shamshiri, V.~Dworak, C.~Weltzien, and R.~Fern{\'a}ndez, ``Soft gripper for small fruits harvesting and pick and place operations,'' \emph{Frontiers in Robotics and AI}, vol.~10, 2023.

\bibitem{visentin2023soft}
F.~Visentin, F.~Castellini, and R.~Muradore, ``A soft, sensorized gripper for delicate harvesting of small fruits,'' \emph{Computers and Electronics in Agriculture}, vol. 213, p. 108202, 2023.

\bibitem{wang2023development}
X.~Wang, H.~Kang, H.~Zhou, W.~Au, M.~Y. Wang, and C.~Chen, ``Development and evaluation of a robust soft robotic gripper for apple harvesting,'' \emph{Computers and Electronics in Agriculture}, vol. 204, p. 107552, 2023.

\bibitem{de2011design}
Z.~De-An, L.~Jidong, J.~Wei, Z.~Ying, and C.~Yu, ``Design and control of an apple harvesting robot,'' \emph{Biosystems Engineering}, vol. 110, no.~2, pp. 112--122, 2011.

\bibitem{ling2004sensing}
P.~P. Ling, R.~Ehsani, K.~C. Ting, Y.-T. Chi, N.~Ramalingam, M.~H. Klingman, and C.~Draper, ``Sensing and end-effector for a robotic tomato harvester,'' in \emph{ASAE annual meeting}.\hskip 1em plus 0.5em minus 0.4em\relax American Society of Agricultural and Biological Engineers, 2004, p.~1.

\bibitem{lehnert2017autonomous}
C.~Lehnert, A.~English, C.~McCool, A.~W. Tow, and T.~Perez, ``Autonomous sweet pepper harvesting for protected cropping systems,'' \emph{IEEE Robotics and Automation Letters}, vol.~2, no.~2, pp. 872--879, 2017.

\bibitem{navas2021soft}
E.~Navas, R.~Fern{\'a}ndez, D.~Sep{\'u}lveda, M.~Armada, and P.~Gonzalez-de Santos, ``Soft grippers for automatic crop harvesting: A review,'' \emph{Sensors}, vol.~21, no.~8, p. 2689, 2021.

\bibitem{mokria2022volume}
M.~Mokria, A.~Gebrekirstos, H.~Said, K.~Hadgu, N.~Hagazi, W.~Dubale, and A.~Br{\"a}uning, ``Volume estimation models for avocado fruit,'' \emph{Plos One}, vol.~17, no.~2, pp. 1--14, 2022.

\bibitem{dechemi2023robotic}
A.~Dechemi, D.~Chatziparaschis, J.~Chen, M.~Campbell, A.~Shamshirgaran, C.~Mucchiani, A.~Roy-Chowdhury, S.~Carpin, and K.~Karydis, ``Robotic assessment of a crop’s need for watering: Automating a time-consuming task to support sustainable agriculture,'' \emph{IEEE Robotics \& Automation Magazine}, vol.~30, no.~4, pp. 52--67, 2023.

\end{thebibliography}


\begin{thebibliography}{99}








































\end{thebibliography}

\end{document}